%% file: paper.tex
\documentclass[conference]{IEEEtran}
\IEEEoverridecommandlockouts
\usepackage{cite}
\usepackage{amsmath,amssymb,amsfonts}
\usepackage{algorithmic}
\usepackage{graphicx}
\usepackage{textcomp}
\usepackage{xcolor}
\usepackage{multirow,booktabs}
\usepackage{bm}
\def\BibTeX{{\rm B\kern-.05em{\sc i\kern-.025em b}\kern-.08em
    T\kern-.1667em\lower.7ex\hbox{E}\kern-.125emX}}

\DeclareMathOperator*{\argmaxA}{arg\,max}

\begin{document}

%-----------------------------------------------
% Each paper should have 6 to maximum 8 pages, 
% including figures, tables, and references.
%-----------------------------------------------

\title{Information Ranking Using Optimum-Path Forest\\
\thanks{The authors are grateful to Petrobras grant \#2014/00545-0, FAPESP grants \#2013/07375-0, \#2014/12236-1, \#2017/25908-6, \#2018/15597-6, \#2019/07665-4, as well as CNPq grants \#307066/2017-7 and \#427968/2018-6.}
}

%-----------------------------------------------

\author{%
\IEEEauthorblockN{
Nathalia Q. Ascen\c{c}\~ao\IEEEauthorrefmark{1}
Luis C. S. Afonso\IEEEauthorrefmark{2},
Danilo Colombo\IEEEauthorrefmark{3},
Luciano Oliveira\IEEEauthorrefmark{4} and
Jo\~ao P. Papa\IEEEauthorrefmark{1}}

\IEEEauthorblockA{\IEEEauthorrefmark{1}%
      UNESP - Univ. Estadual Paulista, School of Sciences, Bauru, Brazil\\
      Email: \{nathalia.ascencao,joao.papa\}@unesp.br}

\IEEEauthorblockA{\IEEEauthorrefmark{2}%
      UFSCar - Federal University of S\~ao Carlos, Department of Computing, S\~ao Carlos, Brazil\\
      Email: sugi.luis@ufscar.br}

\IEEEauthorblockA{\IEEEauthorrefmark{3}%
      Cenpes, Petr\'oleo Brasileiro S.A., \\
      Email: colombo.danilo@petrobras.com.br}

\IEEEauthorblockA{\IEEEauthorrefmark{4}%
      UFBA - Federal University of Bahia, Salvador, Brazil\\
      Email: lrebouca@ufba.br}
}

%\author{%
%    \IEEEauthorblockN{Nathalia Q. Ascen\c{c}\~ao and Jo\~ao P. Papa}
%    \IEEEauthorblockA{%
%      School of Sciences\\
%	  %UNESP - S\~ao Paulo State University\\
%	  UNESP - Univ. Estadual Paulista\\
%      Bauru - SP, Brazil\\
%      nathalia.ascencao@unesp.br\\joao.papa@unesp.br}
%  \and
%    \IEEEauthorblockN{Luis C. S. Afonso}
%    \IEEEauthorblockA{%
%      Department of Computing\\
%      UFSCar - Federal University of S\~ao Carlos\\
%      S\~ao Carlos - SP, Brazil\\%
%	  sugi.luis@ufscar.br}
%  \and      
%    \IEEEauthorblockN{Danilo Colombo}
%    \IEEEauthorblockA{%
%    \textit{Cenpes}\\
%    \textit{Petr\'oleo Brasileiro S.A.}\\
%    Rio de Janeiro - RJ, Brazil\\
%    colombo.danilo@petrobras.com.br}
%  \and
%    \IEEEauthorblockN{Luciano Almeida}
%    \IEEEauthorblockA{%
%      -----\\
%      UFBA - Federal University of Bahia\\
%      Salvador - BA, Brazil\\%
%	  email@email.br}
%}

\maketitle

%-----------------------------------------------

\begin{abstract}
The task of learning to rank has been widely studied by the machine learning community, mainly due to its use and great importance in information retrieval, data mining, and natural language processing. Therefore, ranking accurately and learning to rank are crucial tasks. Context-Based Information Retrieval systems have been of great importance to reduce the effort of finding relevant data. Such systems have evolved by using machine learning techniques to improve their results, but they are mainly dependent on user feedback. Although information retrieval has been addressed in different works along with classifiers based on Optimum-Path Forest (OPF), these have so far not been applied to the learning to rank task. Therefore, the main contribution of this work is to evaluate classifiers based on Optimum-Path Forest, in such a context. Experiments were performed considering the image retrieval and ranking scenarios, and the performance of OPF-based approaches was compared to the well-known SVM-Rank pairwise technique and a baseline based on distance calculation. The experiments showed competitive results concerning precision and outperformed traditional techniques in terms of computational load.
\end{abstract}

%-----------------------------------------------

\IEEEpeerreviewmaketitle
% Introdução sobre a temática de recuperação de informações e ranking
\input{introduction.tex}

% OPF
\input{opf.tex}

% Proposed approach
\input{proposed.tex}
% Methodology
\input{methodology.tex}

% Experiments
\input{experiments.tex}

% Conclusions
\input{conclusions.tex}

%\section*{Acknowledgment}

%-----------------------------------------------

\bibliographystyle{IEEEtran}
\bibliography{paper}

%-----------------------------------------------

\end{document}

%% file: introduction.tex
\section{Introduction}
\label{s.introduction}

Information Retrieval stands for a field of knowledge that aims to return relevant data given an input query, which can be a multimedia content such as an image, audio, video, or a text-based information~\cite{PEDRONETTE20132350}. Nowadays, the amount of data that has been generated has increased considerably. E-mails and multimedia data are interchanged among millions of users daily, thus contributing to spreading communication and increasing the workload in Internet traffic. Therefore, it is highly desired to handle such amount of data efficiently, i.e., to store and further retrieve the relevant information only.

Images have a crucial role in several fields of research, such as medicine, advertising, education, and entertainment, among others~\cite{YOUSSEF20121358}. Content-based Image Retrieval (CBIR) systems come to reduce the costs of manually retrieving relevant data since it is a laborious and very much time-consuming task. CBIR-driven systems aim at retrieving relevant images from a dataset based on features such as color, shape, and texture, and have been assisting in a broad range of applications~\cite{Iliadis-VirtualTouring-IEEE-2013,Milovanovic-CBIR-kinect-IEEE-2013,Bugatti-MedicineCBIR-Elsevier-2014,Memon-ObjectIdentificationCBIR-IEEE-2015,Demir-RemoteSensingCBIR-IEEE-2015}.

However, CBIR systems face the problem of being pretty much user-dependent, which means that it is not straightforward to learn models that can generalize well for every kind of input query. Such techniques make use of relevance feedback from the user to overcome such dependence. The user indicates the images that are relevant (and non-relevant) to their needs, and the process is repeated until the retrieved data is satisfactory~\cite{Tavares-ActiveLearningOPF-ElsevierScience-2011,Pardede-SemanticGap-AmericanScientific-2017,Rani-ActiveLearning-IJICSE-2017,Alzubi-CBIR-study-Elsevier-2015}. Hence, ranking accurately is crucial for CBIR systems.

In this sense, machine learning techniques have been applied in the context of learning to rank. Younus et al.~\cite{Younus-CBIR-PSO-Kmeans-Springer-2015} used image features such as color histogram, color moment, co-occurrence matrices, and wavelet moment for measuring the similarity among samples. Then, they applied $k$-means with the Particle Swarm Optimization algorithm for retrieving images. Irtaza et al.~\cite{Irtaza-ANN-CBIR-Springer-2014} proposed a method that uses an in-depth texture analysis, a learning scheme based on $k$-nearest neighbors and a neural network to the retrieval task. A Bayesian network based on feedback relevance was applied in~\cite{Satish-CBIR-bayesiannetwork-IEEE-2017} on medical image retrieval using feature analysis such as color, texture, and shape jointly with some visual descriptors. %\textcolor{red}{Mahyoub et al.~\cite{MahyoubICDSE:2018}}

In~\cite{Ningthoujam-ANN-SVM-CBIR-IndianJournal-2017}, an artificial neural network along with Support Vector Machine (SVM) was employed on face image retrieval and recognition. Support Vector Machine also has been applied in content-based image retrieving 
~\cite{Pavani-CBIR-machinelearning-Springer-2014,Wang-SVM-feedbackrelevance-Elsevier-2013,Sugamya-SVM-CBIR-IEEE-2016,Hu-activeleaningSVM-Elsevier-2013,Wang-SVM-feedback-imageretrieval-Elsevier-2015,Saad-CBIR-SVM-histogram-IEEE-2017,Seth-LBPSVM-IJARCS-2017,Yogen-LocalSVM-IRJET-2017,Rao-AnovelFbRelevance-WWW-2018} to reduce the semantic gap problem by enhancing the image classification process; as an active classifier combining classification with active learning based on feedback relevance; together with visual descriptors to improve retrieval performance, and also in deep learning as proposed in~\cite{Mohamed-ConvolutionalCBIR-Springer-2018}, where a framework based on Convolutional Neural Networks and SVM was used on feature extraction, classification, and image retrieval.

A few years ago, the Optimum-Path Forest (OPF) framework was proposed to handle the problem of pattern classification as a graph partitioning task. The framework comprises supervised~\cite{Papa-SupervisedOPF-JohnWiley-2009,Papa-EfficientSupervisedOPF-ElsevierScience-2012,Papa-kNN_OPF-ElsevierScience-2017}, semi-supervised~\cite{Papa-SemiSupervisedOPF-ElsevierScience-2016}, and unsupervised versions \cite{Papa-DataClusteringAsOPF-JohnWiley-2009}. Such approaches work by mapping the classification problem as a graph partition task, where the nodes stand for the samples that are represented by their corresponding feature vectors and are connected through some adjacency relation that has been defined previously. In a nutshell, nodes are classified based on a competitive process among key samples that try to conquer others offering them optimum-path costs. The key samples, also called prototypes, are the ones that best represent different classes and are chosen based on a specific heuristic.

Although OPF has been used in several areas, only a few works have applied such a technique to the context of image retrieval. Tavares et al.~\cite{Tavares-ActiveLearningOPF-ElsevierScience-2011} proposed two approaches for content-based image retrieval based on OPF, where the main idea is to ask the user to mark some relevant images which are further used as prototypes for a new training step. Later on, the remaining data is classified and sorted in such a way that contains only the relevant images, which are presented to the user once more. This process is repeated until the user is satisfied. 

Dhawale and Joglekar~\cite{Dhawale-OPF-CBIR-iPGCON-2015} compared OPF against other classifiers for image retrieval purposes, and concluded that it could be much faster than widely used techniques such as Artificial Neural Networks, Support Vector Machines, and the $k$-Nearest Neighbours ($k$-NN) classifier. However, those works are based on user feedback only. Therefore, the main contribution of our work is to introduce the OPF-Ranking (OPF-R), which is an OPF-based approach that can rank images automatically, i.e., without user intervention. The OPF-R was evaluated under three well-known image datasets where there were considered two relevance metrics. The performance was compared against a baseline technique and the SVM-Rank, in which OPF-R showed competitive results and lower ranking computing times.

The remainder of this paper is organized as follows. Section~\ref{s.OPF} reviews the theoretical background concerning OPF-based classifiers. The proposed approach and methodology are presented in Sections~\ref{s.proposed} and~\ref{s.methodology}, respectively. Section~\ref{s.experiments} discusses the experiments and results. Finally, Section~\ref{s.conclusions} states conclusions and future works.

%% file: opf.tex
\section{Optimum-Path Forest}
\label{s.OPF}

The Optimum-Path Forest is a framework for the design of graph-based classifiers. Let ${\cal Z}$ be a labeled dataset, such that ${\cal Z} = {\cal Z}_1 \cup {\cal Z}_2$, in which ${\cal Z}_1$ and ${\cal Z}_2$ are training and testing sets, respectively. OPF encodes each sample $\bm{u} \in {\cal Z}$ as a graph node, and the graph is initially designed based on a predefined adjacency relation ${\cal A}$ with edges weighted by the distance between the feature vectors of their connecting nodes. 

The general OPF training algorithm is divided into two parts: (i) to find a set of samples called prototypes, and (ii) to compute optimum-path trees (OPTs) rooted at them. The prototypes are the most representative samples from each class. The definition of \emph{most representative sample}, as well as the method for computing prototypes, are different for each variant of OPF, and they are explained in the following sections. 

Let ${\cal P}$ be the set of prototypes such that ${\cal P}\subset{\cal Z}_1$. The OPTs are built through a competitive process in which samples try to ``conquer'' each other by offering costs. The competition starts at the prototypes that offer their best cost to the remaining training samples (i.e., non-prototype samples). The costs are defined by a path-cost function, which is also different for each OPF variant. 

Let $\bm{p}\in{\cal P}$ and $\bm{s}\in{\cal Z}_1\backslash{\cal P}$ be some prototype and non-prototype samples, respectively. Suppose that sample $\bm{s}$ is conquered by a sample $\bm{p}$ that offers to it the best cost. Upon such an assumption, $\bm{p}$ assigns its label to $\bm{s}$, and $\bm{s}$ is added to tree rooted at $\bm{p}$. Notice that prototypes cannot be conquered, and a class is represented by at least one tree. The outcome of the training step is a set of optimum-path trees (i.e., optimum-path forest) ${\cal G}_{tr} = ({\cal Z}_1, {\cal A})$.

Concerning the classification step, a node $\bm{v} \in {\cal Z}_2$ is connected to ${\cal G}_{tr}$ according to ${\cal A}$ (i.e., if ${\cal A}$ is a $k$-NN adjacency relation, then $\bm{v}$ is first connected to the $k$-nearest samples from ${\cal G}_{tr}$). Besides, the sample $\bm{u} \in {\cal Z}_1$ that offers the best cost to $\bm{v}$ assigns its label to it. This work makes use of two variants of the supervised version whose working mechanisms are further explained.

%--------------------------------------------------
\subsection{OPF with Complete Graph (CG-OPF)}
\label{ss.CG-OPF}

This supervised variant was introduced by Papa et al.~\cite{Papa-SupervisedOPF-JohnWiley-2009} and implements the complete graph as the adjacency relation (i.e., all nodes are connected). As aforementioned, the first step in the training phase is to find the set of prototypes ${\cal P}$, which is computed through the minimum spanning tree (MST) over ${\cal G}_{tr}$. The prototype nodes are the samples belonging to the intersection region among classes since they are more likely to be misclassified.

The following step is the competition process, which is carried out using Equations~\ref{eq.OPF.fmax.init} and~\ref{eq.OPF.fmax.prop} that stand for the initialization and propagation of costs, respectively:
\begin{equation}
\label{eq.OPF.fmax.init}
f_{\mathrm{max}}(\langle
\mathbf{p}\rangle) = \left\{ \begin{array}{ll}
  0 & \mbox{if \textbf{p} $\in$ {\cal P},} \\
+\infty & \mbox{otherwise}
\end{array}\right.
\end{equation}
and
\begin{equation}
\label{eq.OPF.fmax.prop}
f_{\mathrm{max}}(\pi_\mathbf{p} \cdot \langle \mathbf{p,v} \rangle) = \max\{f_{\mathrm{max}}(\pi\mathbf{_p}),d(\mathbf{p,v})\} ,
\end{equation}
in which $f$ is a real-valued path-cost function, $\pi_\textbf{p} \cdot \langle\textbf{p, v}\rangle$ stands for the concatenation of path $\pi_\textbf{p}$ (i.e., a sequence of adjacent nodes starting from any node and with terminus at node $\textbf{p}$) with an edge $\langle\textbf{p, v}\rangle$, and $d$ denotes a distance function.

\begin{figure}[!htb]
\centerline{
  \includegraphics[scale=0.3]{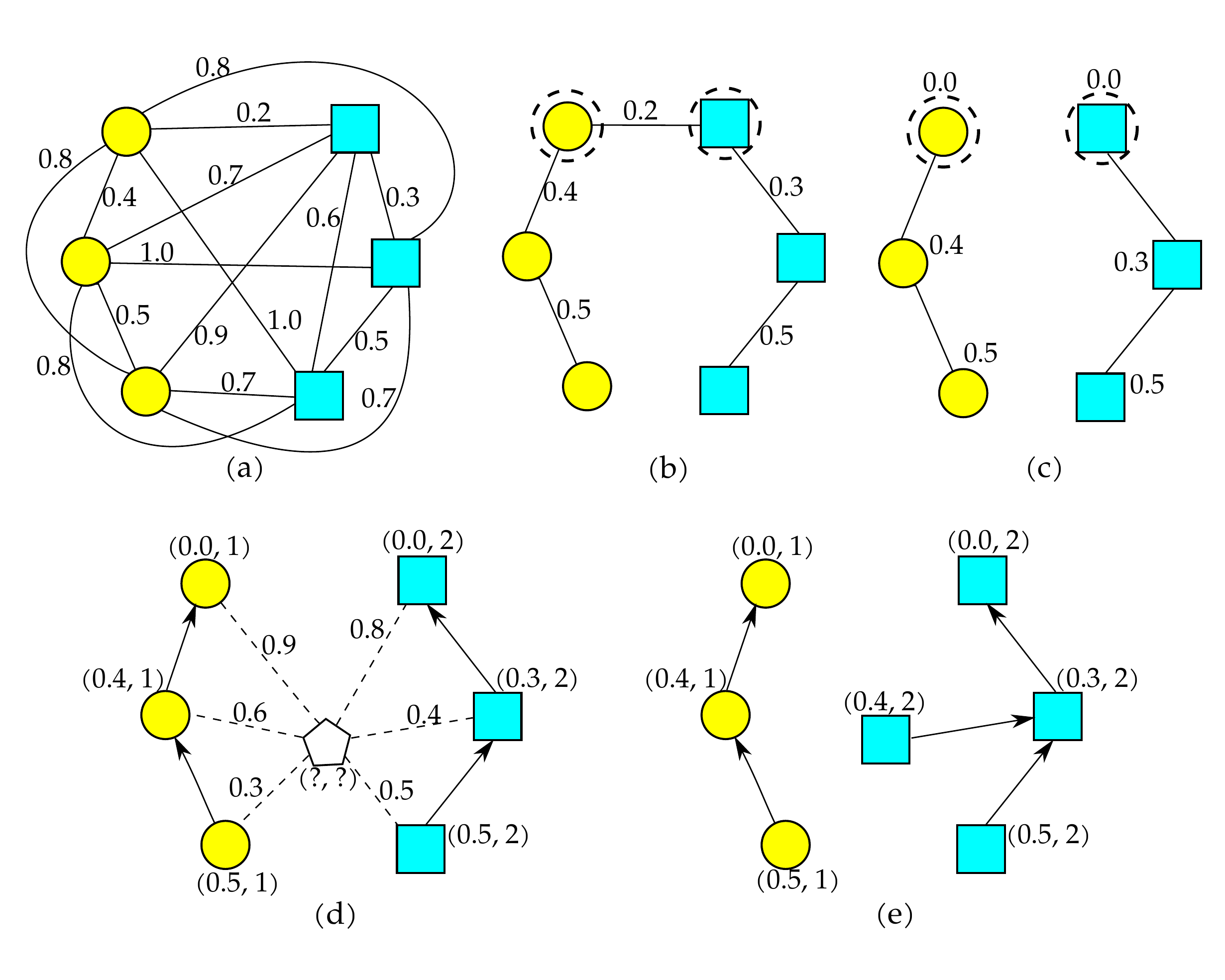}}
\caption{\label{f.opf_cg_full}OPF with complete graph. Training phase: (a) a two-class training graph with weighted arcs, (b) an MST with prototypes highlighted, and (c) optimum-path forest generated during the training phase with costs over the nodes (notice the prototypes have zero cost). Classification phase: (d) sample penthagon is connected to all training nodes, and (e) penthagon is conquered by a sample from the class ``square'', and it receives the ``square"\ label.}
\end{figure}

The conquering of samples happens during the propagation of costs where prototypes offer their optimum-path costs to other samples. The sample $\bm{v}$ is conquered by sample $\bm{p}$ that minimizes (\ref{eq.OPF.fmax.prop}). The classification is based on the connecting force between samples from ${\cal Z}_1$ and testing samples from ${\cal Z}_2$, and works similarly to the training phase. Each $\bm{t} \in{\cal Z}_2$ is connected to ${\cal G}_{tr}$ obbeying ${\cal A}$. Then, it is evaluated the sample $\bm{v}^\ast \in {\cal Z}_1$ that satisfies Equation~\ref{eq.OPF.min-max}:

\begin{equation}
\label{eq.OPF.min-max}
C(\mathbf{t}) = \arg \min \max _{\mathbf{v}\in {\cal Z}_1} \{C(\mathbf{v}),d(\mathbf{v,t})\}, 
\end{equation}
where $C$ is the cost of the sample. Figure~\ref{f.opf_cg_full} depicts the training and classification processes.

%--------------------------------------------------
\subsection{OPF with k-nearest neighbors Graph ($k$-NN-OPF)}
\label{ss.KNN-OPF}

The OPF with k-nearest neighbors Graph was proposed by Papa et al.~\cite{PapaGBRPR:2009,PapaISVC:2008,PapaPRL:2017}, whose main differences to CG-OPF are the adjancency relation, the weighting of nodes, and how prototypes are computed. The $k$-NN-OPF employs the $k$-nearest neighbors as the adjacency relation, and the nodes are now weighted by a probability density function. The method for building the set of prototypes must be changed since a $k$-NN adjacency relation does not guarantee a connected graph. Instead, the set ${\cal P}$ is computed based on region density values where samples of higher density are selected as prototypes. Such approach is similar to selecting the centroids of clusters~\cite{RosaICPR:14}. Hence, $k$-NN-OPF is understood as a ``dual version'' of CG-OPF (minimization problem) since it aims at maximizing the cost of each sample according to Equation~\ref{eq.OPF.max-knn}:

\begin{equation}
\label{eq.OPF.max-knn}
\max f(\pi_\mathbf{u}), \forall \mathbf{u} \in {\cal Z}_1.
\end{equation}

Besides, samples $\bm{u} \in {\cal Z}_1$ are weighted by a function $\rho(u)$ that computes the probability density value as follows:

\begin{eqnarray}
\label{eq.OPF.pdf}
  \rho(\mathbf{u}) & = & \frac{1}{\sqrt{2\pi\sigma^2k}} \sum_{\forall \mathbf{v} \in {{\cal A}_k}(\mathbf{u})} \exp\left(\frac{-d(\mathbf{u,v})}{2\sigma^2}\right),
\end{eqnarray}
where ${\cal A}_{k}(\textbf{u})$ stands for the $k$-nearest neighbors of sample $\textbf{u}$, $d_{\mathrm{max}} = max\{d(\textbf{u, v}) \in G_{tr} \}$, and $\sigma = d_{\mathrm{max}}/3$. After computing the density values for all training samples, $k$-NN-OPF starts the competition process through the path-cost function $f_{\mathrm{min}}$ defined as follows:

\begin{eqnarray}
\label{eq.OPF.min-knn}
f_{\mathrm{min}}(\langle \mathbf{v} \rangle) & = & \left\{ \begin{array}{ll} 
    \rho(\mathbf{v})    & \mbox{if $\textbf{v} \in {\cal P}$} \\
    \rho(\mathbf{v}) - 1     & \mbox{otherwise}
 \end{array}\right. \nonumber \\
f_{\mathrm{min}}(\pi_\mathbf{u}\cdot (\mathbf{u,v})) &=& \min\{f_{\mathrm{min}}(\pi_\mathbf{u}),\rho(\mathbf{v})\}.
\end{eqnarray}
The upper formulation stands for the proper initialization of the training nodes. The term $\rho(\mathbf{v}) - 1 $ must be used to avoid over clustering since a plateau of densities may occur.

The competition itself stands for the propagation of costs among samples. Since the prototype holds the higher cost of its optimum-path tree, the idea is to conquer samples with lower costs. Finally, the sample that maximizes $f_{\mathrm{min}}$ for a given sample $\textbf{v}$ will be the one to conquer it.

The classification of samples in ${\cal Z}_2$ is performed similarly to the conquering process. The first step computes the $k$-nearest neighbors from ${\cal Z}_1$ to a testing sample $t \in {\cal Z}_2$. Finally, it is verified which node $\textbf{s}^\ast\in {\cal Z}_1$ satisfies the equation below:

\begin{equation}
	\label{e.opfknn_classification}
	C(\textbf{t})=\argmaxA_{\textbf{v}\in {\cal Z}_1}\min\{C(\textbf{v}),\rho(\textbf{t})\}.
\end{equation}

\begin{figure}[!htb]
\centerline{
  \includegraphics[scale=0.3]{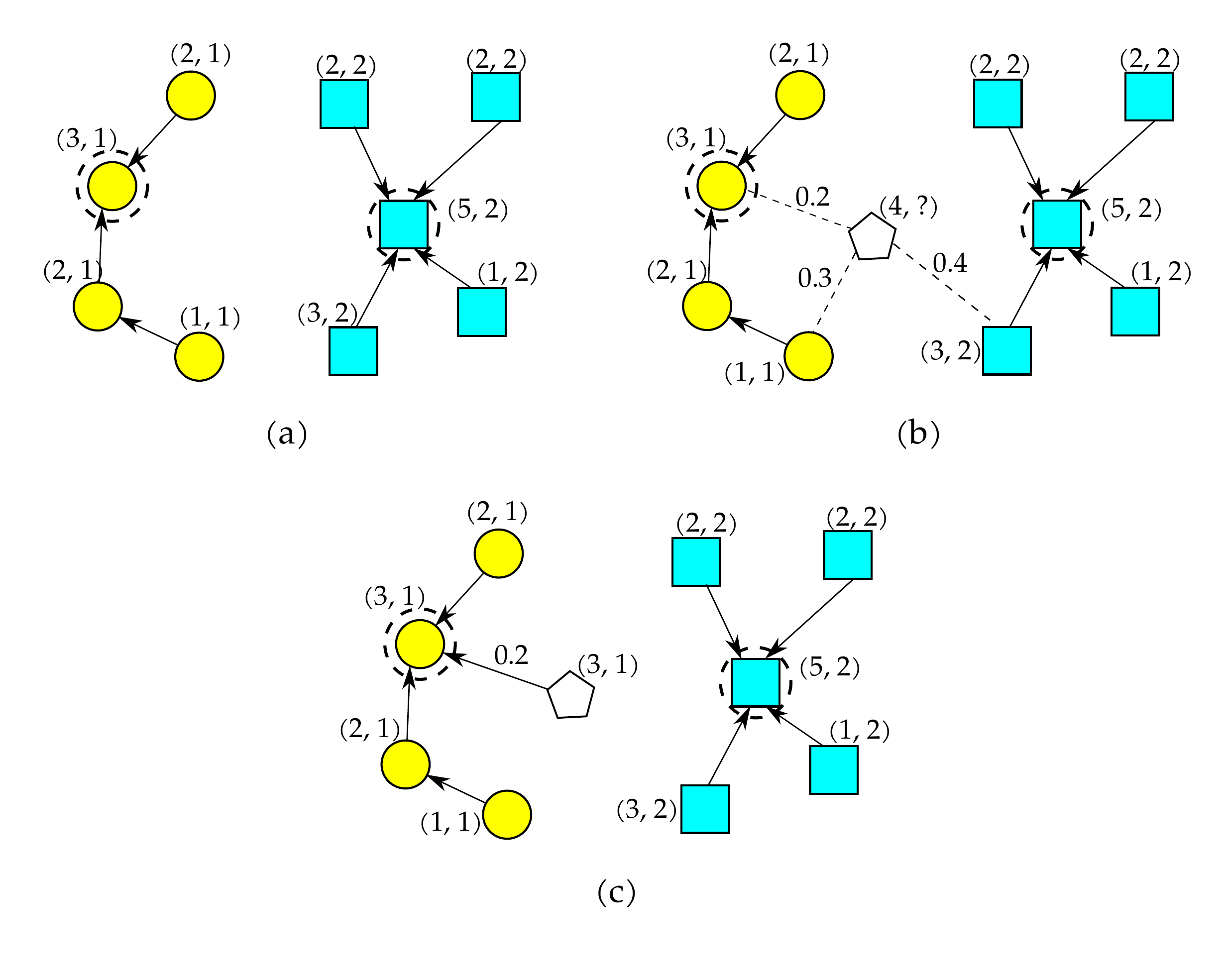}}
\caption{\label{f.opf_knn_classification}OPF with $k$-NN graph classification: (a) optimum-path forest generated during the training phase with the pair (cost, label) over the nodes, (b) sample penthagon is connected to its $k$-nearest training nodes, and (c) penthagon is conquered by a sample from the class ``circle'', and it receives the ``circle"\ label.}
\end{figure}

Figure~\ref{f.opf_knn_classification} depicts the classification process.

%% file: proposed.tex
\section{Proposed Approach}
\label{s.proposed}

The idea of ranking in CBIR is to retrieve the most similar images for a given query. However, OPF does not rely on any similarity metric solely. Instead, the ranking problem is designed as a connectivity problem, where a testing sample is assigned to a class based on the strength of its connection to each sample in ${\cal G}_{tr}$, where the path cost defines the strength level (i.e., the better the path-cost, the stronger is the strength). Therefore, (OPF-R) maps the similarity as the strength of the query to the samples from the trained model (i.e., the stronger is the connection between a query and a model sample, the higher is their similarity).

After training OPF-R, a given query $q$ is further connected to ${\cal G}_{tr}$ according to the adjacency relation ${\cal A}$, i.e., if ${\cal A}$ is the complete graph type, then $q$ is connected to all samples in ${\cal A}$; if ${\cal A}$ is the $k$-NN graph type, then $q$ is connected to the $k$-nearest samples in ${\cal A}$. Notice that the training is perfomed as described in Section~\ref{s.OPF}.

\begin{figure}[!htb]
\centerline{
  \includegraphics[scale=0.35]{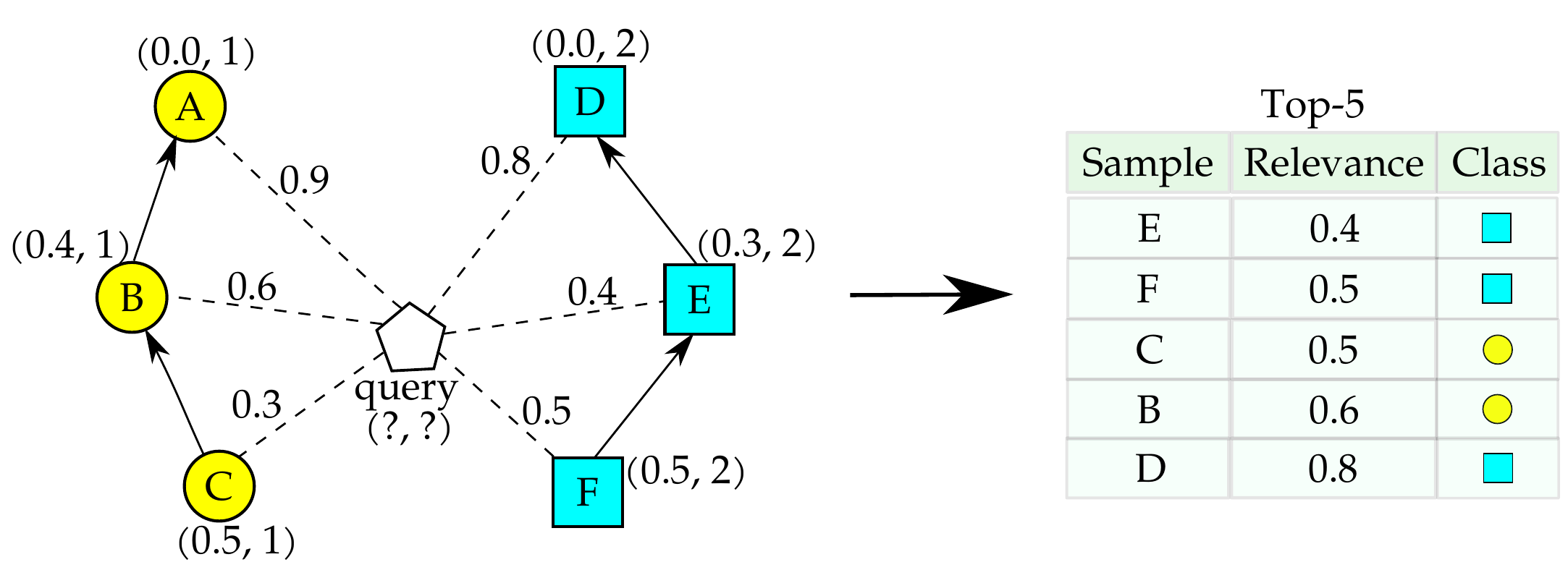}}
\caption{\label{f.workflow}Ranking by OPF. The query sample is connected to the training set according and costs are further computed. Samples of lower cost are the most strongly connected ones.}
\end{figure}

The next step is to perform the competition process, similarly to the classification process. The difference is that both CG-OPF and $k$-NN-OPF store all path-costs offered to $q$. In the end, the path-costs are sorted, and the $r$ best costs comprise the ranking list. Figure~\ref{f.workflow} depicts the ranking approach through OPF-R with a complete graph (CG-OPF-R) considering the Top-5 results. One interesting point is that OPF ranks samples natively, i.e., there is no need to perform complex changes in the original classifier to support both ranking and retrieval.

%% file: methodology.tex
\section{Methodology}
\label{s.methodology}

This section presents the data and techniques employed to validate the proposed approach.

%--------------------------------------------------
\subsection{Datasets}
\label{ss.datasets}

The proposed approach was evaluated in three datasets with detailed information presented in Table~\ref{t.datasets}. Notice that Brodatz dataset has originally a total of $112$ images, and was later expanded to $1,792$ images by dividing the original ones into $16$ parts. This strategy was applied to increase the number of samples per class.

\begin{table}[h]
\centering
\caption{Dataset information.}
\label{t.datasets}
\resizebox{\columnwidth}{!}{%
\begin{tabular}{ccccc}
\hline
\textbf{dataset}    &    \textbf{type}  & \textbf{\# images} & \textbf{\# classes} & \textbf{images per classes} \\ \hline
   Brodatz          &       texture     &      $1,792$       &         $112$       &           $16$          \\
   Caltech101       &   objects/scenes  &      $8,677$       &         $101$       &         $40-800$        \\
   MPEG-7           &       shape       &      $1,500$       &         $70$        &           $20$          \\\hline
\end{tabular}}
\end{table}

%--------------------------------------------------
\subsection{Features}
\label{ss.features}

The experiments considered a total of seven features. However, they are not applied to all datasets since their nature is not suitable for all cases. For instance, Brodatz is a dataset characterized by texture images. Therefore, only Local Binary Pattern (LBP)~\cite{02-LBP} and Statistical Analysis of Structural Information (SASI)~\cite{03-SASI} features are computed. Table~\ref{t.features} presents the seven extracted features.

\begin{table}[h]
\centering
\caption{Feature information.}
\label{t.features}
\resizebox{\columnwidth}{!}{%
\begin{tabular}{ccc}
\hline
\textbf{feature}                                                        &    \textbf{type}  & \textbf{dataset} \\ \hline
 Auto-color correlation (ACC)~\cite{Huang-1997}                         &       color       &    Caltech101    \\
 Border/Interior Pixel Classification (BIC)~\cite{Stehling-2002}        &       color       &    Caltech101    \\
 Color Coherence Vectors (CCV)~\cite{97-CCV}                            &       color       &    Caltech101    \\
 Local Color Histogram (LCH)~\cite{91-LCH}                              &       color       &    Caltech101    \\
 Local Binary Pattern (LBP)~\cite{02-LBP}                               &      texture      &     Brodatz      \\
 Statistical Analysis of Structural Information (SASI)~\cite{03-SASI}   &      texture      &     Brodatz      \\
 Spherical Pyramid-Technique (SPYTEC)~\cite{07-SPYTEC}                  &       shape       &     MPEG-7       \\ \hline
\end{tabular}}
\end{table}

%--------------------------------------------------
\subsection{Experimental Setup}
\label{ss.setup}

The experiments were carried out using three configurations of training and testing set (querying set) sizes: (i) $25\%$ samples for training and $75\%$ samples for classification, (ii) $50\%$ samples for training and $50\%$ samples for classification, and (iii) $75\%$ samples for training and $25\%$ samples for classification. The sets were randomly generated. Table~\ref{t.parameters} presents the parameter values used in this work\footnote{These parameters were obtained empirically.}.

%Regarding the parameters, CG-OPF is parameterless whereas $k$NN-OPF had its parameter $k$ set as $--$.

\begin{table}[h]
\centering
\caption{Parameters.}
\label{t.parameters}
\begin{tabular}{ccc}
\hline
\textbf{technique}  &    \textbf{parameters}        \\ \hline
     CG-OPF         &               --              \\
     $k$-NN-OPF      &      $k_{\mathrm{max}} = 20$          \\
     SVM-Rank       & $c=0.05$, $l=0$, $e=0.001$    \\ \hline
\end{tabular}
\end{table}
\noindent
The remaining parameters of SVM-Rank were set in their default values as defined in~\cite{Joachims2006:SIGKDD}.

%--------------------------------------------------
\subsection{Relevance Metric}
\label{ss.relevance}

The accuracy is measured by two metrics as described:

\begin{itemize}
    \item Normalized Discounted Cumulative Gain (NDCG): The NDCG computes the cumulative gain (i.e., the sum of the relevance score of the candidate samples) considering the ordering in the ranking. The metric is based on the Discounted Cumulative Gain (DCG) that reduces the score of relevant data in a logarithmical proportional to their position in the ranking, Equation~\ref{eq.dcg}. The lower the position, the higher is the penalization:
    \begin{equation}
        \label{eq.dcg}
        DCG_r = \sum^{r}_{i=1}\frac{2^{{\mathrm{rel}}_i}-1}{{\mathrm{log}}_2(i+1)},
    \end{equation}
where $rel_i$ is the graded relevance of the result at position $i$. Thus, NDCG is defined as follows:
    \begin{equation}
        \label{eq.ndcg}
        NDCG_r = \frac{DCG_r}{DCG_{r_{\mathrm{ideal}}}},
    \end{equation}
where $DCG_{r_{\mathrm{ideal}}}$ stands for the case where the data is sorted by their relevance score.
    \item Mean Average Precision (MAP): The MAP is commonly used to evaluate ranking methods applied to binary score problems and computes the mean average precision for each query as follows:
    \begin{equation}
        MAP = \frac{1}{n}\sum^{n}_{q=1}AP_q,
    \end{equation}
where $n$ is the number of queries and AP is the average precision of a single query and defined as:
    \begin{equation}
        AP = \frac{\sum^{n}_{r=1}(P@r\times I(rel_{r}==1))}{\sum^{n}_{r=1}I(rel_{r}==1)},
    \end{equation}
being P@r the precision:
    \begin{equation}
        P@r = \frac{1}{r}\sum^{r}_{i=1}(rel_{r}==1),
    \end{equation}
which is the normalized number of relevant candidates in the first $r$ positions.
\end{itemize}

Both metrics use the absolute relevance (i.e., a label of $0$ or $1$ is assigned to the candidate based on its relevance). The relevance of a candidate is defined by comparing its label to the query after the ranking. The candidate is assigned a relevance $0$ if it does not have the same label as the query, or $1$ if the candidate and query share the same label.

%--------------------------------------------------

%% file: experiments.tex
\section{Experimental Results}
\label{s.experiments}

The experimental results are organized by datasets with tables presenting the NDGC and MAP values for each scenario (i.e., dataset $\times$ training/testing sets configuration $\times$ descriptor). Besides evaluating OPF-Rank in a great variety of cases, we also compared its performance against a distance-based technique (Distance) and the well-known SVM-Rank considering the top-$10$, top-$15$, and top-$20$ rankings. Notice that the Distance technique computes the ranking based on the distance function suggested by the author of the descriptor (i.e., each descriptor has its most appropriate distance function). The best results (i.e., for each configuration, descriptor and top-r) are shown in bold, according to the Wilcoxon signed-rank test with the significance of $0.05$.  Besides, a hold-out approach with $10$ runs with randomly generated training and testing sets was applied for validation purposes.

%--------------------------------------------------
\subsection{Brodatz}
\label{sss.brodatz}

The experimental results are presented in Table~\ref{t.brodatz2575}. One can observe that the best results achieved by CG-OPF and $k$-NN-OPF were using the SASI descriptor regardless of the training/testing sets configuration. Concerning the comparison between the OPF-based approaches, CG-OPF obtained a better performance over $k$-NN-OPF in the $25\%\times75\%$ configuration, whereas the performance was very close in the remaining configurations. Among all techniques, SVM-Rank presented the best overall results.

\begin{table}[h]
\caption{\label{t.brodatz2575}Results concerning Brodatz dataset.}%using the configuration $25\%\times75\%$.}
\resizebox{\columnwidth}{!}{%
\begin{tabular}{cccccccc}
\hline
\multicolumn{8}{c}{\textbf{$25\%\times75\%$}} \\
\hline
\multirow{3}{*}{\textbf{technique}} & \multirow{3}{*}{\textbf{descriptor}} & \multicolumn{6}{c}{\textbf{top-r}} \\ 
\cmidrule(r){3-8} 
&  & \multicolumn{2}{c}{\textbf{10}} & \multicolumn{2}{c}{\textbf{15}} & \multicolumn{2}{c}{\textbf{20}} \\ 
\cmidrule(r){3-4} \cmidrule(r){5-6} \cmidrule(r){7-8} 
&  & NDGC & MAP & NDGC & MAP & NDGC & MAP \\ \hline
                 Distance              & LBP   & \textbf{0.370}  & \textbf{0.225}  & 0.377  & \textbf{0.182}  & 0.382  & \textbf{0.152}      \\
                                       & SASI  & 0.392          & 0.260         & 0.393  & 0.170  & 0.396  & 0.140      \\
                 CG-OPF                & LBP   & \textbf{0.354}  & 0.162         & 0.371  & 0.124  & 0.375  & 0.102      \\
                                       & SASI  & 0.405          & 0.199         & 0.410  & 0.154  & 0.411  & 0.127      \\
                 $k$-NN-OPF             & LBP   & 0.328          & 0.135         & 0.345  & 0.106  & 0.349  & 0.087      \\
                                       & SASI  & 0.378          & 0.189         & 0.381  & 0.146  & 0.383  & 0.120      \\
                 SVM-Rank              & LBP   & 0.362         & 0.145      & \textbf{0.383} & 0.115 & \textbf{0.388} & 0.094      \\
                                       & SASI  & \textbf{0.431} & \textbf{0.328} & \textbf{0.434} & \textbf{0.264} & \textbf{0.436} & \textbf{0.225} \\ \hline
%\end{tabular}}
%\end{table}

%-------------------------
\multicolumn{8}{c}{\textbf{$50\%\times50\%$}} \\
%\begin{table}[h]
%\caption{\label{t.brodatz5050}Results using the configuration $50\%\times50\%$.}
%\resizebox{\columnwidth}{!}{%
%\begin{tabular}{cccccccc}
\hline
\multirow{3}{*}{\textbf{technique}} & \multirow{3}{*}{\textbf{descriptor}} & \multicolumn{6}{c}{\textbf{top-r}} \\ 
\cmidrule(r){3-8} 
&  & \multicolumn{2}{c}{\textbf{10}} & \multicolumn{2}{c}{\textbf{15}} & \multicolumn{2}{c}{\textbf{20}} \\ 
\cmidrule(r){3-4} \cmidrule(r){5-6} \cmidrule(r){7-8} 
&  & NDGC & MAP & NDGC & MAP & NDGC & MAP \\ \hline
Distance        & LBP  & 0.368  & 0.202  & 0.378  & 0.164  & 0.386 & 0.140  \\     
                & SASI & 0.406  & 0.289  & 0.411  & 0.236  & 0.411 & 0.201  \\ %\cline{2-8} 
CG-OPF          & LBP  & 0.391  & 0.214  & 0.402  & 0.174  & 0.410 & 0.148  \\                                     
                & SASI & 0.428  & 0.222  & 0.431  & 0.176  & 0.432 & 0.148  \\ %\cline{2-8}  
$k$-NN-OPF       & LBP  & 0.361  & 0.197  & 0.370  & 0.160  & 0.378 & 0.137  \\                                     
                & SASI & 0.432  & 0.308  & 0.435  & 0.251  & 0.437 & 0.212  \\ %\cline{2-8} 
SVM-Rank        & LBP  & \textbf{0.409} & \textbf{0.224} & \textbf{0.420} & \textbf{0.182} & \textbf{0.429} & \textbf{0.155}  \\                                     
                & SASI & \textbf{0.451} & \textbf{0.321} & \textbf{0.455} & \textbf{0.262} & \textbf{0.457} & \textbf{0.222} \\ \hline
%\end{tabular}}
%\end{table}

%-------------------------
\multicolumn{8}{c}{\textbf{$75\%\times25\%$}} \\
%\begin{table}[h]
%\caption{\label{t.brodatz7525}Results using the configuration $75\%\times25\%$.}
%\resizebox{\columnwidth}{!}{%
%\begin{tabular}{cccccccc}
\hline
\multirow{3}{*}{\textbf{technique}} & \multirow{3}{*}{\textbf{descriptor}} & \multicolumn{6}{c}{\textbf{top-r}} \\ 
\cmidrule(r){3-8} 
&  & \multicolumn{2}{c}{\textbf{10}} & \multicolumn{2}{c}{\textbf{15}} & \multicolumn{2}{c}{\textbf{20}} \\ 
\cmidrule(r){3-4} \cmidrule(r){5-6} \cmidrule(r){7-8} 
&  & NDGC & MAP & NDGC & MAP & NDGC & MAP \\ \hline
Distance    & LBP   & 0.341  & 0.145  & 0.359  & 0.118  & 0.364 & 0.099 \\     
            & SASI  & 0.406  & 0.289  & 0.411  & 0.236  & 0.411 & 0.201 \\ %\cline{2-8} 
CG-OPF      & LBP   & 0.410  & 0.255  & 0.418  & 0.210  & 0.423 & 0.179 \\                                     
            & SASI  & 0.402  & 0.209  & 0.405  & 0.166  & 0.407 & 0.139 \\ %\cline{2-8}  
$k$-NN-OPF   & LBP   & 0.378  & 0.235  & 0.385  & 0.194  & 0.389 & 0.165 \\                                     
            & SASI  & 0.403  & 0.334  & 0.403  & 0.274  & 0.406 & 0.236 \\ %\cline{2-8}
SVM-Rank    & LBP   & \textbf{0.429} & \textbf{0.267} & \textbf{0.437} & \textbf{0.220} & \textbf{0.442} & \textbf{0.187} \\
            & SASI  & \textbf{0.457} & \textbf{0.379} & \textbf{0.457} & \textbf{0.311} & \textbf{0.460} & \textbf{0.268} \\ \hline
\end{tabular}}
\end{table}

%--------------------------------------------------
\subsection{Caltech101}
\label{sss.caltech}

The results are presented in Table~\ref{t.caltech2575}, in which LCH and BIC provided the best results. CG-OPF showed better relevance values over $k$-NN-OPF in all considered configurations. Except for Distance, all techniques are benefited by increasing the number of training samples. The best overall results were achieved by SVM-Rank.

\begin{table}[h]
\caption{\label{t.caltech2575}Results concerning Caltech101 dataset.}% using the configuration $25\%\times75\%$.}
\resizebox{\columnwidth}{!}{%
\begin{tabular}{cccccccc}
\hline
\multicolumn{8}{c}{\textbf{$25\%\times75\%$}} \\
\hline
\multirow{3}{*}{\textbf{technique}} & \multirow{3}{*}{\textbf{descriptor}} & \multicolumn{6}{c}{\textbf{top-r}} \\ 
\cmidrule(r){3-8} 
&  & \multicolumn{2}{c}{\textbf{10}} & \multicolumn{2}{c}{\textbf{15}} & \multicolumn{2}{c}{\textbf{20}} \\ 
\cmidrule(r){3-4} \cmidrule(r){5-6} \cmidrule(r){7-8} 
&  & NDGC & MAP & NDGC & MAP & NDGC & MAP \\ \hline
Distance        
& ACC                                 
& 0.252 & \textbf{0.151} & 0.270 & \textbf{0.143} & 0.283 & \textbf{0.136}             
\\                                     
& BIC                                 
& 0.264 & \textbf{0.162} & 0.281 & \textbf{0.155} & 0.294 & \textbf{0.149}             
\\                                    
& CCV                                 
& 0.242 & \textbf{0.140} & 0.263 & \textbf{0.132} & 0.277 & \textbf{0.126}              
\\                                     
& LCH                                 
& 0.269 & 0.162 & 0.289 & 0.151 & 0.302 & 0.142             
\\ %\cline{2-8} 
CG-OPF           
& ACC                                 
& 0.250 & 0.141 & 0.274 & 0.132 & 0.286 & 0.125 
\\                                     
& BIC                                 
& 0.259 & 0.151 & \textbf{0.284} & 0.142 & 0.267 & 0.135            
\\                                     
& CCV                                 
& 0.238 & 0.125 & 0.260 & 0.118 & 0.276 & 0.112 
\\                                     
& LCH                                 
& 0.264 & 0.156 & 0.282 & 0.144 & 0.299 & 0.138             
\\ %\cline{2-8} 
$k$-NN-OPF         
& ACC                                
& 0.232 & 0.132 & 0.254 & 0.124 & 0.266 & 0.116
\\                                     
& BIC                                 
& 0.241 & 0.142 & 0.259 & 0.133 & 0.273 & 0.126             
\\                                     
& CCV                                 
& 0.222 & 0.118 & 0.242 & 0.111 & 0.257 & 0.105
\\                                     
& LCH                                 
& 0.246 & 0.146 & 0.262 & 0.136 & 0.277 & 0.128             
\\ %\cline{2-8} 
SVM-Rank 
& ACC                                 
& \textbf{0.260} & 0.146 & \textbf{0.258} & 0.137 & \textbf{0.298} & 0.129               
\\                                     
& BIC                                 
& \textbf{0.270} & 0.157 & \textbf{0.263} & 0.147 & \textbf{0.306} & 0.140            
\\                                     
& CCV                                
& \textbf{0.248} & 0.130 & \textbf{0.244} & 0.122 & \textbf{0.288} & 0.116
\\                                     
& LCH                                 
& \textbf{0.275} & \textbf{0.174} & \textbf{0.294} & \textbf{0.164} & \textbf{0.311} & \textbf{0.156}            
\\ \hline
%\end{tabular}}
%\end{table}

%-------------------------
\multicolumn{8}{c}{\textbf{$50\%\times50\%$}} \\
%\begin{table}[h]
%\caption{\label{t.caltech5050}Results using the configuration $50\%\times50\%$.}
%\resizebox{\columnwidth}{!}{%
%\begin{tabular}{cccccccc}
\hline
\multirow{3}{*}{\textbf{technique}} & \multirow{3}{*}{\textbf{descriptor}} & \multicolumn{6}{c}{\textbf{top-r}} \\ 
\cmidrule(r){3-8} 
&  & \multicolumn{2}{c}{\textbf{10}} & \multicolumn{2}{c}{\textbf{15}} & \multicolumn{2}{c}{\textbf{20}} \\ 
\cmidrule(r){3-4} \cmidrule(r){5-6} \cmidrule(r){7-8} 
&  & NDGC & MAP & NDGC & MAP & NDGC & MAP \\ \hline
Distance        
& ACC                                 
& 0.283 & 0.182 & 0.301 & 0.174 & 0.314 & \textbf{0.167}              
\\                                     
& BIC                                 
& 0.295 & \textbf{0.186} & 0.312 & 0.186 & 0.325 & \textbf{0.180}             
\\                                    
& CCV                                 
& 0.273 & 0.171 & 0.294 & 0.163 & 0.309 & \textbf{0.157}              
\\                                     
& LCH                                 
& 0.300 & 0.193 & 0.320 & 0.182 & 0.333 & 0.173            
\\ %\cline{2-8} 
CG-OPF           
& ACC                                 
& 0.281 & 0.172 & 0.305 & 0.163 & 0.317 & 0.157 
\\                                     
& BIC                                 
& 0.290 & 0.182 & 0.310 & 0.173 & 0.325 & \textbf{0.175}           
\\                                     
& CCV                                 
& 0.269 & 0.156 & 0.291 & 0.149 & 0.307 & 0.143 
\\                                     
& LCH                                 
& 0.295 & 0.187 & 0.313 & 0.175 & 0.330 & 0.172             
\\ %\cline{2-8} 
$k$-NN-OPF         
& ACC                                
& 0.263 & 0.163 & 0.285 & 0.155 & 0.297 & 0.147
\\                                     
& BIC                                 
& 0.272 & 0.173 & 0.290 & 0.164 & 0.304 & 0.157             
\\                                     
& CCV                                 
& 0.253 & 0.149 & 0.273 & 0.142 & 0.287 & 0.137
\\                                     
& LCH                                 
& 0.277 & 0.177 & 0.293 & 0.167 & 0.310 & 0.159             
\\ %\cline{2-8} 
SVM-Rank 
& ACC                                 
& \textbf{0.291} & \textbf{0.177} & \textbf{0.316} & \textbf{0.169} & \textbf{0.329} & \textbf{0.160}               
\\                                     
& BIC                                 
& \textbf{0.301} & \textbf{0.190} & \textbf{0.321} & \textbf{0.180} & \textbf{0.337} & \textbf{0.171}            
\\                                     
& CCV                                
& \textbf{0.264} & \textbf{0.146} & \textbf{0.287} & \textbf{0.138} & \textbf{0.304} & \textbf{0.132}
\\                                     
& LCH                                 
& \textbf{0.306} & \textbf{0.205} & \textbf{0.325} & \textbf{0.195} & \textbf{0.342} & \textbf{0.187}  
\\ \hline
%\end{tabular}}
%\end{table}

%-------------------------
\multicolumn{8}{c}{\textbf{$75\%\times25\%$}} \\
%\begin{table}[h]
%\caption{\label{t.caltech7525}Results using the configuration $75\%\times25\%$.}
%\resizebox{\columnwidth}{!}{%
%\begin{tabular}{cccccccc}
\hline
\multirow{3}{*}{\textbf{technique}} & \multirow{3}{*}{\textbf{descriptor}} & \multicolumn{6}{c}{\textbf{top-r}} \\ 
\cmidrule(r){3-8} 
&  & \multicolumn{2}{c}{\textbf{10}} & \multicolumn{2}{c}{\textbf{15}} & \multicolumn{2}{c}{\textbf{20}} \\ 
\cmidrule(r){3-4} \cmidrule(r){5-6} \cmidrule(r){7-8} 
&  & NDGC & MAP & NDGC & MAP & NDGC & MAP \\ \hline
Distance        
& ACC                                 
& 0.340 & 0.239 & 0.358 & 0.231 & 0.371 & 0.224              
\\                                     
& BIC                                 
& 0.352 & 0.241 & 0.369 & 0.243 & 0.382 & 0.237             
\\                                    
& CCV                                 
& 0.330 & 0.228 & 0.351 & 0.221 & 0.366 & 0.214              
\\                                     
& LCH                                 
& 0.357 & 0.250 & 0.377 & 0.239 & 0.390 & 0.230            
\\ %\cline{2-8} 
CG-OPF           
& ACC                                 
& 0.338 & 0.229 & 0.362 & 0.220 & 0.374 & 0.213
\\                                     
& BIC                                 
& 0.347 & 0.239 & 0.367 & 0.230 & 0.382 & 0.223           
\\                                     
& CCV                                 
& 0.326 & 0.213 & 0.348 & 0.206 & 0.364 & 0.201 
\\                                     
& LCH                                 
& 0.352 & \textbf{0.248} & \textbf{0.372} & 0.232 & 0.387 & 0.230             
\\ %\cline{2-8} 
$k$-NN-OPF         
& ACC                                
& 0.321 & 0.220 & 0.342 & 0.212 & 0.354 & 0.204
\\                                     
& BIC                                 
& 0.329 & 0.230 & 0.347 & 0.221 & 0.361 & 0.216             
\\                                     
& CCV                                 
& 0.310 & 0.206 & 0.330 & 0.199 & 0.338 & 0.193
\\                                     
& LCH                                 
& 0.336 & 0.235 & 0.351 & 0.224 & 0.367 & 0.216           
\\ %\cline{2-8} 
SVM-Rank 
& ACC                                 
& \textbf{0.348} & \textbf{0.234} & \textbf{0.373} & \textbf{0.226} & \textbf{0.386} & \textbf{0.217}              
\\                                     
& BIC                                 
& \textbf{0.358} & \textbf{0.247} & \textbf{0.378} & \textbf{0.237} & \textbf{0.396} & \textbf{0.228}            
\\                                     
& CCV                                
& \textbf{0.336} & \textbf{0.218} & \textbf{0.359} & \textbf{0.210} & \textbf{0.376} & \textbf{0.205}
\\                                     
& LCH                                 
& \textbf{0.363} & \textbf{0.262} & \textbf{0.382} & \textbf{0.252} & \textbf{0.399} & \textbf{0.253}  
\\ \hline
\end{tabular}}
\end{table}

%--------------------------------------------------
\subsection{MPEG-7}
\label{sss.1mpeg}

The results are presented in Table~\ref{t.mpeg2575}. Once again, CG-OPF outperformed $k$-NN-OPF in all configurations. An interesting behavior is that changing the configuration $25\%\times75\%$ to $50\%\times50\%$ provided a very small increase in the results, whereas the configuration $75\%\times25\%$ caused a drop in the results. In this dataset, CG-OPF showed competitive results when compared to SVM-Rank.

\begin{table}[h]
\caption{\label{t.mpeg2575}Results concerning MPEG-7 dataset.}% using the configuration $25\%\times75\%$ e base de imagens MPEG-7.}
\resizebox{\columnwidth}{!}{%
\begin{tabular}{cccccccc}
\hline
\multicolumn{8}{c}{\textbf{$25\%\times75\%$}} \\
\hline
\multirow{3}{*}{\textbf{technique}} & \multirow{3}{*}{\textbf{descriptor}} & \multicolumn{6}{c}{\textbf{top-r}} \\ 
\cmidrule(r){3-8} 
&  & \multicolumn{2}{c}{\textbf{10}} & \multicolumn{2}{c}{\textbf{15}} & \multicolumn{2}{c}{\textbf{20}} \\ 
\cmidrule(r){3-4} \cmidrule(r){5-6} \cmidrule(r){7-8} 
&  & NDGC & MAP & NDGC & MAP & NDGC & MAP \\ \hline
Distance & \multirow{4}{*}{SPYTEC} &0.061 & 0.030 & 0.071 & 0.027 & 0.074 & 0.026 \\                                  
CG-OPF   & & \textbf{0.076} & 0.078 & 0.071 & 0.030 & 0.088 & 0.029 \\                                      
$k$-NN-OPF& & 0.066 & 0.031 & 0.072 & 0.028 & 0.082 & 0.027\\                                      
\emph{SVM-Rank} & & 0.075 & \textbf{0.034} & \textbf{0.082} & \textbf{0.031} & \textbf{0.092} & \textbf{0.031} \\ \hline
%\end{tabular}}
%\end{table}

%-------------------------
\multicolumn{8}{c}{\textbf{$50\%\times50\%$}} \\
%\begin{table}[h]
%\caption{\label{t.mpeg5050}Results using the configuration $50\%\times50\%$ e base de imagens MPEG-7.}
%\resizebox{\columnwidth}{!}{%
%\begin{tabular}{cccccccc}
\hline
\multirow{3}{*}{\textbf{technique}} & \multirow{3}{*}{\textbf{descriptor}} & \multicolumn{6}{c}{\textbf{top-r}} \\ 
\cmidrule(r){3-8} 
&  & \multicolumn{2}{c}{\textbf{10}} & \multicolumn{2}{c}{\textbf{15}} & \multicolumn{2}{c}{\textbf{20}} \\ 
\cmidrule(r){3-4} \cmidrule(r){5-6} \cmidrule(r){7-8} 
&  & NDGC & MAP & NDGC & MAP & NDGC & MAP \\ \hline
Distance&\multirow{4}{*}{SPYTEC}& 0.087 & 0.051 & 0.093 & 0.051 & 0.108 & 0.052         
\\                                    
CG-OPF && 0.094 & 0.054 & 0.098 & 0.053 & 0.113 & 0.055        
\\                                      
$k$-NN-OPF&& 0.086 & 0.050 & 0.093 & 0.050 & 0.107 & 0.052        
\\                                     
\emph{SVM-Rank} && \textbf{0.095} &  \textbf{0.055} &  \textbf{0.101} &  \textbf{0.056} &  \textbf{0.118} &  \textbf{0.057}         
\\ \hline
%\end{tabular}}
%\end{table}

%-------------------------
\multicolumn{8}{c}{\textbf{$75\%\times25\%$}} \\
%\begin{table}[h]
%\caption{\label{t.mpeg7525}Results using the configuration $75\%\times25\%$ e base de imagens MPEG-7.}
%\resizebox{\columnwidth}{!}{%
%\begin{tabular}{cccccccc}
\hline
\multirow{3}{*}{\textbf{technique}} & \multirow{3}{*}{\textbf{descriptor}} & \multicolumn{6}{c}{\textbf{top-r}} \\ 
\cmidrule(r){3-8} 
&  & \multicolumn{2}{c}{\textbf{10}} & \multicolumn{2}{c}{\textbf{15}} & \multicolumn{2}{c}{\textbf{20}} \\ 
\cmidrule(r){3-4} \cmidrule(r){5-6} \cmidrule(r){7-8} 
&  & NDGC & MAP & NDGC & MAP & NDGC & MAP \\ \hline
Distance  & \multirow{4}{*}{SPYTEC} & 0.134 & 0.097 & 0.140 & 0.095 & 0.147 & 0.095 \\                                     
CG-OPF    &  & 0.130 & 0.103 & 0.141 & 0.101 & 0.145 & \textbf{0.102} \\                                       
$k$-NN-OPF &  & 0.126 & 0.099 & 0.136 & 0.098 & 0.138 & 0.098 \\                                       
\emph{SVM-Rank}   &  & \textbf{0.137} & \textbf{0.103} & \textbf{0.145} & \textbf{0.102} & \textbf{0.150} & \textbf{0.103} \\ \hline
\end{tabular}}
\end{table}

\subsection{Discussion}
\label{ss.discussion}

In this section, we present a discussion concerning the results obtained in the experiments. Besides, we also provided an additional study concerning the computational load as well. Tables~\ref{t.brodatz_comp},~\ref{t.caltech_comp}, and~\ref{t.mpeg_comp} present the results concerning Brodatz, Caltech101, and MPEG-7 datasets, respectively. The results stand for the average (seconds) ranking time over $10$ runs. As one can observe, OPF-based approaches are pretty much faster than Distance and SVM-Rank techniques (e.g., $1.8$ to $2.2$ times faster). However, SVM-Rank achieved the best results in most cases but with up $4\%$ of superiority over OPF-based techniques.

If one takes into account the trade-off between ranking relevance and retrieving time, OPF-based approaches figure as the most prominent ones, since they achieved results close to the SVM-Rank, but faster. Another point that should be highlighted is that the proposed approach was little modified to handle ranking problems; meanwhile, SVM-Rank needed a considerable adaptation in its working mechanism. In other others, we expect to achieve better results with a more in-depth change in OPF-based competition process to better adapt to ranking-driven applications.

Also, we are not taking into account the training time, which is supposed to be even faster concerning OPF-based approaches. Particular attention is given to CG-OPF, which does not comprise any parameter beforehand, thus turning out to be easier to be set up and with no need for a fine-tuning step. Additionally, although $k$-NN-OPF figures one parameter, its training time is faster than SVM-Rank, in which the number of parameters depends on the kernel function used.

Based on the above assumptions, we conclude that OPF-based classifiers are suitable for ranking purposes, even with its native version. We expect that better results may come with more in-depth modifications that shall affect little the efficiency of the methods.

\begin{table}[h]
\caption{\label{t.brodatz_comp}Computational load [s] concerning Brodatz dataset.}
\resizebox{\columnwidth}{!}{%
\begin{tabular}{ccccccccccc}
\hline
\multirow{3}{*}{\textbf{technique}} & \multirow{3}{*}{\textbf{descriptor}} & \multicolumn{9}{c}{\textbf{top-r}} \\ 
\cmidrule(r){3-11} 
&  & \multicolumn{3}{c}{\textbf{10}} & \multicolumn{3}{c}{\textbf{15}} & \multicolumn{3}{c}{\textbf{20}} \\ 
\cmidrule(r){3-5} \cmidrule(r){6-8} \cmidrule(r){9-11} 
&                                      & 25x75 & 50x50 & 75x25 & 25x75 & 50x50 & 75x25 & 25x75 & 50x50 & 75x25 \\ \hline
                 Distance              & LBP   &  31.01 & 33.01  &  32.14 & 32.00  & 35.00  & 32.37 & 33.06  & 36.83  & 32.95      \\
                                       & SASI  &  31.49 & 34.65  &  33.33 & 33.26  &  35.39 & 34.03 & 34.51  & 37.41  &  34.11   \\
                 CG-OPF                & LBP   & 19.15  & \textbf{18.23}  &  \textbf{16.22} & 20.50  & \textbf{19.45}  & \textbf{16.47}   & 21.15 & \textbf{19.90}  &  \textbf{17.33}    \\
                                       & SASI  & \textbf{16.74}  & \textbf{18.85}  &  \textbf{16.64} & \textbf{17.90}  & \textbf{19.95}  & \textbf{16.75}   & \textbf{18.20}  & \textbf{20.20}  & \textbf{17.95} \\
                 $k$-NN-OPF             & LBP   &  \textbf{17.29} & 19.34  &  17.24 & \textbf{18.40}  & 20.45  & 17.57   & \textbf{18.73}  & 20.74  & 18.32 \\
                                       & SASI  &  18.46 & 20.45  &  18.37 & 19.61  & 21.61  & 18.55   & 20.12  & 22.06  & 19.18 \\
                 SVM-Rank              & LBP   &  32.35 & 34.35  &  34.50 & 33.36  & 35.36  & 35.01   & 34.05  & 35.97  & 35.21 \\
                                       & SASI  &  33.12 & 35.12  &  36.19 & 34.49  & 36.50  & 36.49   & 34.87  & 36.83  & 37.11 \\ \hline
\end{tabular}}
\end{table}

\begin{table}[h]
\caption{\label{t.caltech_comp}Computational load [s] concerning Caltech101 dataset.}
\resizebox{\columnwidth}{!}{%
\begin{tabular}{ccccccccccc}
\hline
\multirow{3}{*}{\textbf{technique}} & \multirow{3}{*}{\textbf{descriptor}} & \multicolumn{9}{c}{\textbf{top-r}} \\ 
\cmidrule(r){3-11} 
&  & \multicolumn{3}{c}{\textbf{10}} & \multicolumn{3}{c}{\textbf{15}} & \multicolumn{3}{c}{\textbf{20}} \\ 
\cmidrule(r){3-5} \cmidrule(r){6-8} \cmidrule(r){9-11} 
            &       & 25x75 & 50x50 & 75x25 & 25x75 & 50x50 & 75x25 & 25x75 & 50x50 & 75x25 \\ \hline
Distance    & ACC   &  40.98 & 40.16 & 39.25 & 41.72 & 40.95 & 40.09 & 42.22 & 41.38 & 40.49 \\                                     
            & BIC   &  40.66 & 40.29 & 39.35 & 41.06 & 41.48 & 40.71 & 41.62 & 41.81 & 40.91 \\                                    
            & CCV   &  41.42 & 39.16 & 37.66 & 41.74 & 39.48 & 37.98 & 42.19 & 39.69 & 38.09 \\                                     
            & LCH   &  39.46 & 38.07 & 36.87 & 40.53 & 38.10 & 36.60 & 40.64 & 38.84 & 37.38 \\ %\cline{2-8} 
CG-OPF      & ACC   &  \textbf{21.03} & \textbf{21.71} & \textbf{20.66} & \textbf{21.86} & \textbf{22.12} & \textbf{20.98} & \textbf{22.55} & \textbf{22.52} & \textbf{21.44} \\                                     
            & BIC   &  \textbf{22.31} & \textbf{22.60} & \textbf{21.77} & \textbf{23.03} & \textbf{23.06} & \textbf{21.98} & \textbf{23.40} & \textbf{23.54} & \textbf{22.43} \\                                    
            & CCV   &  \textbf{21.30} & \textbf{21.71} & \textbf{20.77} & \textbf{22.06} & \textbf{22.29} & \textbf{21.32} & \textbf{22.73} & \textbf{23.09} & \textbf{22.20} \\                                     
            & LCH   &  \textbf{22.55} & \textbf{22.82} & \textbf{21.76} & \textbf{14.09} & \textbf{23.10} & \textbf{21.97} & \textbf{23.38} & \textbf{23.46} & \textbf{22.32} \\ %\cline{2-8} 
$k$-NN-OPF   & ACC   &  22.62 & 22.69 & 21.55 & 13.96 & 23.12 & 21.97 & 14.26 & 23.49 & 22.35 \\                                     
            & BIC   &  23.45 & 24.05 & 22.21 & 23.95 & 24.38 & 22.55 & 24.39 & \textbf{24.79} & 22.98 \\                                    
            & CCV   &  22.27 & 23.52 & 21.66 & \textbf{14.10} & 24.11 & 22.35 & 23.71 & 24.51 & 22.76 \\                                     
            & LCH   &  \textbf{23.13} & 25.29 & 23.32 & \textbf{23.25} & 25.86 & 23.88 & \textbf{24.13} & 26.14 & 24.21 \\ %\cline{2-8} 
SVM-Rank    & ACC   &  41.86 & 43.08 & 41.55 & 42.392 & 43.192 & 42.495 & 43.012 & 44.672 & 43.297 \\                                     
            & BIC   &  42.39 & 41.87 & 40.32 & 42.94 & 42.47 & 40.97 & 43.41 & 43.03 & 41.44 \\                                    
            & CCV   &  43.50 & 43.18 & 41.89 & 43.93 & 43.61 & 42.35 & 44.27 & 43.98 & 42.77 \\                                     
            & LCH   &  42.61 & 43.93 & 41.44 & 43.64 & 45.06 & 42.77 & 44.30 & 45.37 & 43.09 \\ \hline
\end{tabular}}
\end{table}

\begin{table}[h]
\caption{\label{t.mpeg_comp}Computational load [s] concerning MPEG-7 dataset.}
\resizebox{\columnwidth}{!}{%
\begin{tabular}{ccccccccccc}
\hline
\multirow{3}{*}{\textbf{technique}} & \multirow{3}{*}{\textbf{descriptor}} & \multicolumn{9}{c}{\textbf{top-r}} \\ 
\cmidrule(r){3-11} 
&  & \multicolumn{3}{c}{\textbf{10}} & \multicolumn{3}{c}{\textbf{15}} & \multicolumn{3}{c}{\textbf{20}} \\ 
\cmidrule(r){3-5} \cmidrule(r){6-8} \cmidrule(r){9-11} 
            &       & 25x75 & 50x50 & 75x25 & 25x75 & 50x50 & 75x25 & 25x75 & 50x50 & 75x25 \\ \hline
\multirow{4}{*}{SPYTEC} & Distance  & 28.18  & 28.17  & 29.66 & 29.35 & 29.58 & 30.86 & 29.75 & 29.80 & 31.12 \\                                     
                        & CG-OPF    & \textbf{18.35}  & \textbf{18.11}  & \textbf{19.35} & \textbf{20.23} & \textbf{20.27} & \textbf{21.76} & \textbf{21.55} & \textbf{21.72} & \textbf{22.61} \\                                       
                        & $k$-NN-OPF & 20.11  & 20.34  & 21.32 & 21.24 & 21.24 & 21.95 & 22.00 & 22.50 & 24.17 \\                                     
                        & SVM-Rank  & 32.78  & 33.13  & 34.66 & 33.55 & 33.81 & 34.63 & 34.12 & 34.59 & 35.46 \\ \hline
\end{tabular}}
\end{table}

%% file: conclusions.tex
\section{Conclusions}
\label{s.conclusions}

In this work, we introduced two OPF variants to the context of content-based image retrieval and ranking. Both approaches, i.e., CG-OPF and $k$-NN-OPF, achieved promising results when compared to SVM-Rank, but being faster for ranking purposes. Although the latter one figured as the most accurate technique in almost all simulations, the best trade-off between effectiveness and efficiency was achieved by OPF.

As future works, we intend to change the OPF working mechanism and adapt some parts to handle better the problem of image ranking, as well as to consider different distance functions for the arc-weights.